\documentclass[10pt,twocolumn,letterpaper]{article}

\usepackage[pagenumbers]{cvpr} % To force page numbers, e.g. for an arXiv version

\usepackage{graphicx}
\usepackage{amsmath}
\usepackage{amssymb}
\usepackage{booktabs}
\usepackage[para]{footmisc}
\usepackage{enumitem}

\usepackage[pagebackref,breaklinks,colorlinks]{hyperref}

\usepackage[capitalize]{cleveref}
\crefname{section}{Sec.}{Secs.}
\Crefname{section}{Section}{Sections}
\Crefname{table}{Table}{Tables}
\crefname{table}{Tab.}{Tabs.}

\def\cvprPaperID{21} % *** Enter the CVPR Paper ID here
\def\confName{EPIC and Ego4D Workshop at CVPR 22}
\def\confYear{2022}

\newcommand{\dataset}{EPIC-KITCHENS Text Subset }

\begin{document}

%%%%%%%%% TITLE - PLEASE UPDATE
\title{An Evaluation of OCR on Egocentric Data}

\author{Valentin Popescu\\
{\tt\small bk18529@bristol.ac.uk}
% For a paper whose authors are all at the same institution,
% omit the following lines up until the closing ``}''.
% Additional authors and addresses can be added with ``\and'',
% just like the second author.
% To save space, use either the email address or home page, not both
% \and
% Second Author\\
% Institution2\\
% First line of institution2 address\\
% {\tt\small secondauthor@i2.org}
\and
Dima Damen\\
{\tt\small dima.damen@bristol.ac.uk}
\and
Toby Perrett\\
{\tt\small toby.perrett@bristol.ac.uk}
}

% \author{{Valentine Popescu \hspace{10mm} Dima Damen \hspace{10mm} Toby Perrett} \\
% Department of Computer Science, University of Bristol, UK \\ {\tt\small <first>.<last>@bristol.ac.uk}
% }

\maketitle

\begin{abstract}
In this paper, we evaluate state-of-the-art OCR methods on Egocentric data. We annotate text in EPIC-KITCHENS images, and demonstrate that existing OCR methods struggle with rotated text, which is frequently observed on objects being handled. We introduce a simple rotate-and-merge procedure which can be applied to pre-trained OCR models that halves the normalized edit distance error. This suggests that future OCR attempts should incorporate rotation into model design and training procedures.
\end{abstract}

%%%%%%%%% BODY TEXT - ENTER YOUR RESPONSE BELOW
\section{Introduction}

Optical Character Recognition (OCR) has been deployed in a wide range of environments, from document digitization \cite{doc_digitisation} to road sign recognition \cite{traffic_sign}. Egocentric data
presents a number of additional challenges, such as occlusion, motion blur, orientation and text size, particularly when objects are being handled \cite{epic}.

% Optical Character Recognition (OCR) has improved significantly in recent years, due to the increased availability of data \cite{avail_data}, improved model architectures \cite{alexnet} and larger computational resources \cite{comp_res}.
% This has made it suitable for a wide range of real-world problems, such as document digitisation \cite{doc_digitisation}, road sign recognition\cite{traffic_sign} and industrial inspection.

% However, these environments tend to be relatively clean, with footage taken from static cameras or vision sensors.

% Egocentric data presents a number of additional challenges for OCR methods to overcome, including long, fluid video streams, lack of active curation and interpretation of objects in a human context \cite{ego4d}. 

\begin{figure}[t]
  \centering
  \includegraphics[width=\linewidth]{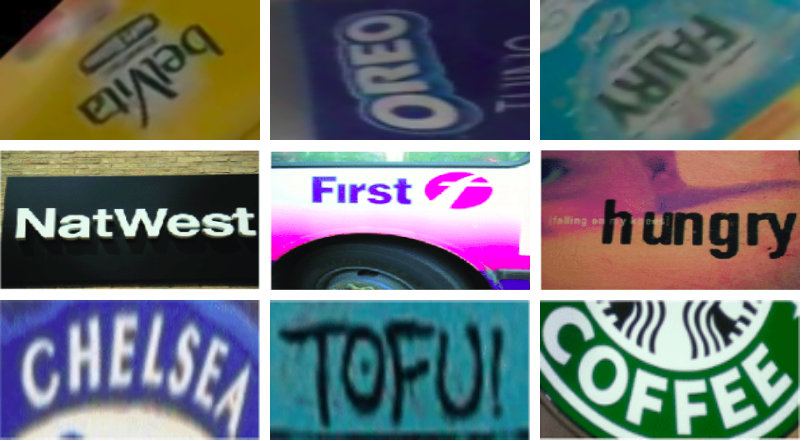}
  \caption{Text crops from the EPIC-KITCHENS \cite{epic} (top), compared to ICDAR 2015 \cite{icdar15} (middle) and SynthText \cite{synthadd} (bottom).}
  \label{fig:samples}
\end{figure}

In this short paper, we provide an initial evaluation of OCR on EPIC-KITCHENS data. In particular, we show that existing pre-trained models struggle with rotated text, which is common in real-life data, but missing from widely-used OCR datasets for model training as demonstrated in Fig. \ref{fig:samples}. We show that significant improvements can be obtained by rotating input frames and merging OCR results from multiple orientations.  Note that this is not intended to be a definitive OCR method - the simplicity highlights a significant oversight in current state-of-the-art OCR methods, training datasets and procedures, suggesting directions for future research.

In summary, our main contributions are:
\begin{itemize}[leftmargin=*,itemsep=-1ex,partopsep=1ex,parsep=1ex]
% \begin{itemize}
    \item We annotate a subset of the EPIC-KITCHENS dataset to provide a ground truth for testing OCR on egocentric data.
    \item We perform an evaluation of pre-trained state-of-the-art OCR methods on this dataset.
    \item We investigate the effect of rotation on OCR accuracy, and propose a simple rotate-and-merge procedure which \emph{halves} the character-normalised edit distance error.
\end{itemize}
In Sec. \ref{sec:background}, we give a brief overview of OCR methods and datasets. In Sec. \ref{sec:dataset}, we introduce the \dataset dataset. In Sec. \ref{sec:method} we introduce the rotate-and-merge procedure, and present results in Sec. \ref{sec:experiments}.

\section{Background}
\label{sec:background}

\begin{figure*}[t]
    \centering
    \includegraphics[width=\textwidth]{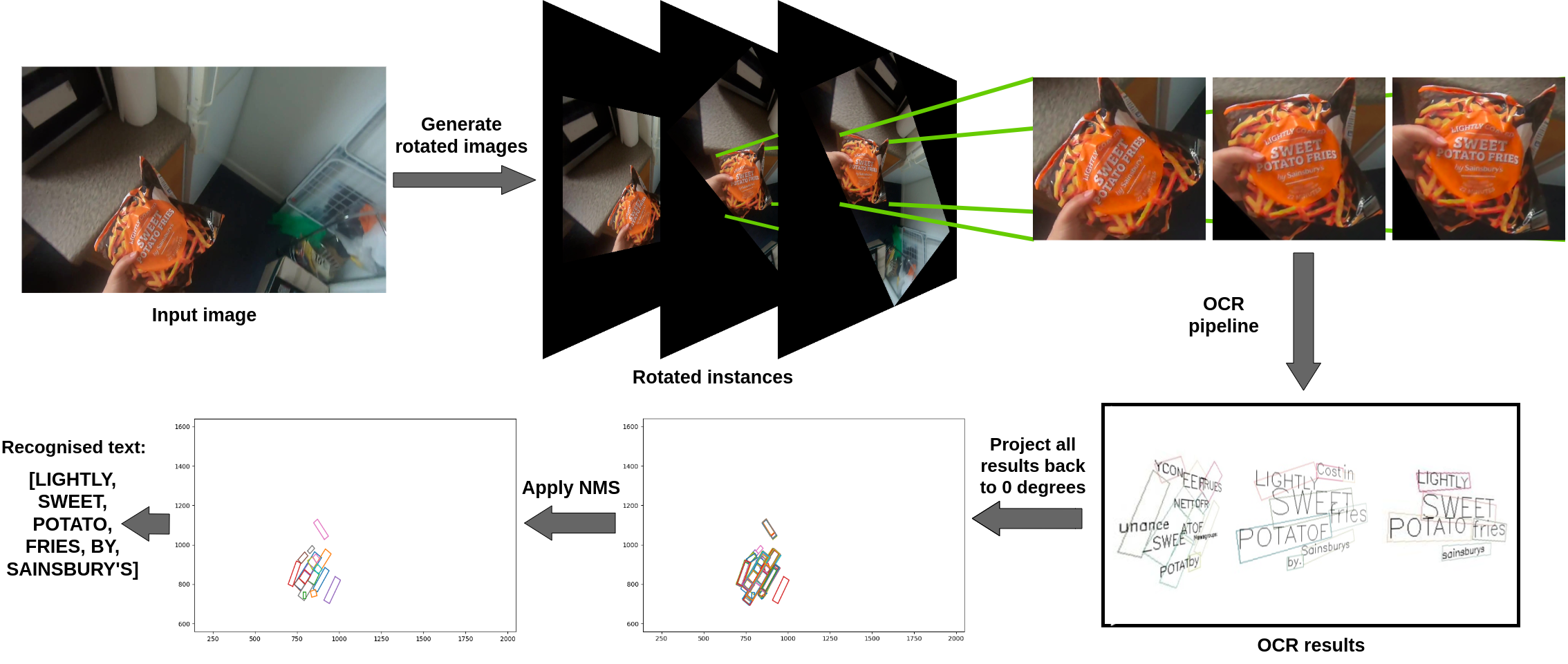}
    \caption{Overview of the proposed rotate-and-merge procedure, applied to an image from EPIC-KITCHENS \cite{epic} using a pre-trained OCR model.}
    \label{fig:method}
    \vspace{-3mm}
\end{figure*}

This section presents a brief overview of OCR methods and datasets.

\noindent\textbf{OCR methods: }
OCR methods typically consist of two stages \cite{mmocr}. First, a detector finds text regions, which are cropped. Second, a recognition model extracts text from the cropped text region, effectively `reading' the text. 
% An example of this process on a frame from EPIC-KITCHENS is shown in Fig. \ref{fig:pipeline}.

Text detectors tend to follow the standard object detection paradigms. Some use single shot detection, similar to SSD \cite{ssd}. A number of them use region proposals similar to the RCNN family of object detectors~\cite{rcnn_family}. An advantage of region proposals is that they can handle a larger range of text sizes, which is especially important for egocentric data.
In this work we use PSENet~\cite{psenet}, which additionally exploits multiple scales (making it well suited to small text) as the detector due to its strong results on a number of baselines \cite{icdar15,ctw}.

\noindent\textbf{Related Efforts: } Some attempts have been made at making recognition robust to rotation, through encouraging robust features \cite{aon} or generating rotated proposals for objects \cite{rcnn_family} and text \cite{rotated_proposals_text}, but these were not evaluated on the extreme rotations and distortions found in egocentric datasets (e.g. CT80 \cite{ct80} and SVTP \cite{svtp}), and their base architectures are outperformed significantly by more recent approaches \cite{sar,abi}.

% The orientation issue was addressed before by Zhanzhan Cheng et al.  or Chiang and Knoblock \cite{recog_mo_ms_ct}, but their work is evaluated on standard benchmark datasets (CT80 \cite{ct80}, SVTP \cite{svtp}) and raster maps respectively, both differing significantly from our egocentric input data. 

\noindent\textbf{OCR datasets: }
% The text detection method was pre-trained on IC15 \cite{icdar15} as it contains incidental scene text. The text recognition method is pre-trained on a combination of SynthText \cite{synthtext}, Syn90k \cite{syn90k}, ICDAR 2015 \cite{icdar15_r}, SVTP \cite{svtp}, CT80 \cite{ct80} and COCO-Text \cite{coco} to ensure more variability. 
OCR datasets tend to focus on near-horizontal text \cite{icdar15}. They come from diverse sources such as street view images \cite{svtp} and synthetic generation \cite{synthadd}. Some have distortions \cite{ctw}, and varying text sizes \cite{ct80}, but none have the amount of rotation and distortion found in naturally-collected egocentric data. Methods are frequently trained on combinations of these datasets to improve generalisation performance \cite{mmocr}.

%------------------------------------------------------------------------
\section{The \dataset}
\label{sec:dataset}
% {\bf Make sure to update the paper title and paper ID in the appropriate place in the tex file.}

% \begin{figure}[t]
%   \centering
%   \includegraphics[width=\linewidth]{latex/images/dataset_examples_9_abstr.png}
%   \caption{Samples from the EPIC-Text-Rotation dataset. The box contours show the results returned by the selected detector.}
%   \label{fig:epic_samples}
% \end{figure}

To evaluate state-of-the-art recognition models and asses their suitability for indoor egocentric data, we curate the \dataset dataset. It contains 490 text cropped images\footnote{Note that whilst 490 images is not suitable for training, it is sufficient to evaluate pre-trained models and demonstrate the effect of rotation.} obtained from EPIC-KITCHENS-100 videos \cite{epic}. The crops were selected based on manual frame selection followed by detection confidence thresholding (from multiple rotations), ensuring varied examples are selected for labelling. 
Each detected text block was labelled by 5 Amazon Mechanical Turk Workers. A consensus was obtained by majority voting on the entire text content. Cases of multiple candidates were resolved manually.
% Samples from dataset are shown in Fig. \ref{fig:epic_samples}.

The dataset contains 120 distinct word instances, and features a long-tail. Only 30\% of the dataset contains image crops oriented around the horizontal ($\pm 15^{\circ}$), showing that a significant proportion of text is not horizontal.
%A total of 42 images were not labelled as workers could not read the text presented, highlighting the complexity of the task.

%-------------------------------------------------------------------------

\section{Proposed Method}\label{sec:method}
\vspace{-1mm}
An initial inspection of OCR outputs from EPIC-KITCHENS images suggests that many of the failed recognitions are on text which is not horizontal, and that the majority of text is not horizontal. To address this, we introduce a simple procedure which exploits pre-trained OCR models. The same input image is rotated, forming a multi-rotation stack of images. These are all then passed through an OCR model individually. Their predictions are then rotated back to horizontal, to be directly comparable, and combined to produce one final output. This is shown in Fig. \ref{fig:method}. We now discuss the method in detail.

We first define a set of rotations every $r^{\circ}$ as $\mathcal{R} = \{ir: i = 0 ... 360/r\}$, and their inverses as $\mathcal{R}^{-1}$. 
We perform inference using a pre-trained OCR model $\Phi$, which takes in an input image and returns a text bounding box $b$ along with the recognised text $t$, bounding box confidence score $\mu$ and text recognition confidence score $\nu$ for each block of text it finds. We denote this $\Phi: R_i(x) \rightarrow \{ (b, t, \mu, \nu)_{ij} : j=1..J_i \}$ where $J_i$ is the number of discovered text blocks for the image $x$ after it rotated by $R \in \mathcal{R}$.

We apply $\Phi$ to all rotated images (so it acts on rotated text), which gives the set of all text blocks from all rotated images. We then apply the inverse rotations to the bounding boxes, so they align with the original, un-rotated image.
$B = \{R^{-1}\Phi(R(x)) : \forall R \in \mathcal{R}\}$ is the set of all text blocks from all rotations, mapped back to the original image.

Finally, we apply the Non-Maximum Suppression algorithm \cite{rcnn_family} to $B$, which takes in a set of boxes and confidence scores and returns the final set of detections for the image. We calculate the confidence score $c_{ij}$ for each box $b_{ij}$ as $c_{ij} = (\mu_{ij} + \nu_{ij}) / 2$.

\section{Experiments}\label{sec:experiments}

In this section, we evaluate a set of OCR methods with and without the rotation pipeline introduced in Sec. \ref{sec:method}.

\noindent\textbf{Dataset:}
We perform all evaluations on the \dataset images, introduced in Sec. \ref{sec:dataset}.

\noindent\textbf{OCR methods:}
PSENet\cite{psenet}, pre-trained on IC15~\cite{icdar15} is used as the detector due to state-of-the-art results on the IC15 benchmark.
We perform experiments using 8 recognition methods from the MMOCR toolbox \cite{mmocr}: ABINet \cite{abi}, SAR \cite{sar}, SATRN \cite{satrn}, RobustScanner \cite{robustscanner}, NRTR \cite{nrtr}, SegOCR \cite{mmocr}, TPS \cite{tps} and CRNN\cite{crnn}. 
These are pre-trained using a combination of 8 datasets: ICDAR11 \cite{icdar11}, ICDAR13 \cite{icdar13}, ICDAR15 \cite{icdar15}, COCO-text \cite{coco}, IIIT5K \cite{iiit5k}, SynthText \cite{synthadd}, SynthAdd \cite{synthadd}, Syn90k \cite{syn90k}. 
These are all evaluated as part of the rotation pipeline (Sec. \ref{sec:method}), as well as baseline versions on un-rotated crops from the detector.

\noindent\textbf{Implementation details:}
We set $r=15^{\circ}$ as a balance between performance and computation cost. All input images are $1920\times1080$. Following standard practice, cropped text regions are resized to $250\times140$ before being passed to the recognition model.

\noindent\textbf{Metrics:}
We evaluate using the following three metrics:
\begin{itemize}[leftmargin=*,itemsep=0ex,partopsep=1ex,parsep=1ex] \vspace{-2mm}
    \item Accuracy: Correct word instances / all word instances.
    \item Average edit distance (Avg. ED): Edit distance per sample, averaged over all samples.
    \item Normalised edit distance (Norm. ED): Edit distance per sample, divided by sample ground truth length, averaged over all samples. 
\end{itemize}
\vspace{-1mm
}
\begin{table}[t]
\centering
\resizebox{0.47\textwidth}{!}{%
\begin{tabular}{c|ccc}
\hline
Method        & Accuracy $\uparrow$ & Avg. ED $\downarrow$ & Norm. ED $\downarrow$ \\
\hline
ABINet \cite{abi}       & 16.5    & 4.53                  & 0.67                            \\
SAR \cite{sar}          & 11.4    & 5.46                  & 0.74                            \\
SATRN \cite{satrn}        & 10.8    & 5.13                  & 0.74                            \\
RobustScanner \cite{robustscanner} & 8.1     & 5.29                  & 0.73                            \\
SegOCR \cite{mmocr}          & 5.1      & 5.54                  & 0.72                          \\
NRTR \cite{nrtr}          & 3.0     & 5.72                  & 0.82                            \\
TPS \cite{tps}          & 2.4     & 5.64                  & 0.84                           \\
CRNN \cite{crnn}          & 1.4     & 5.72                  & 0.87                           \\
\hline
\end{tabular}%
}\caption{Recognition results obtained without the rotation pipeline. ($\uparrow$: higher is better, $\downarrow$: lower is better).}
\label{tab:results_no_rot}
\vspace{-2mm}
\end{table}

\begin{figure}[t]
  \centering
  \includegraphics[width=\linewidth]{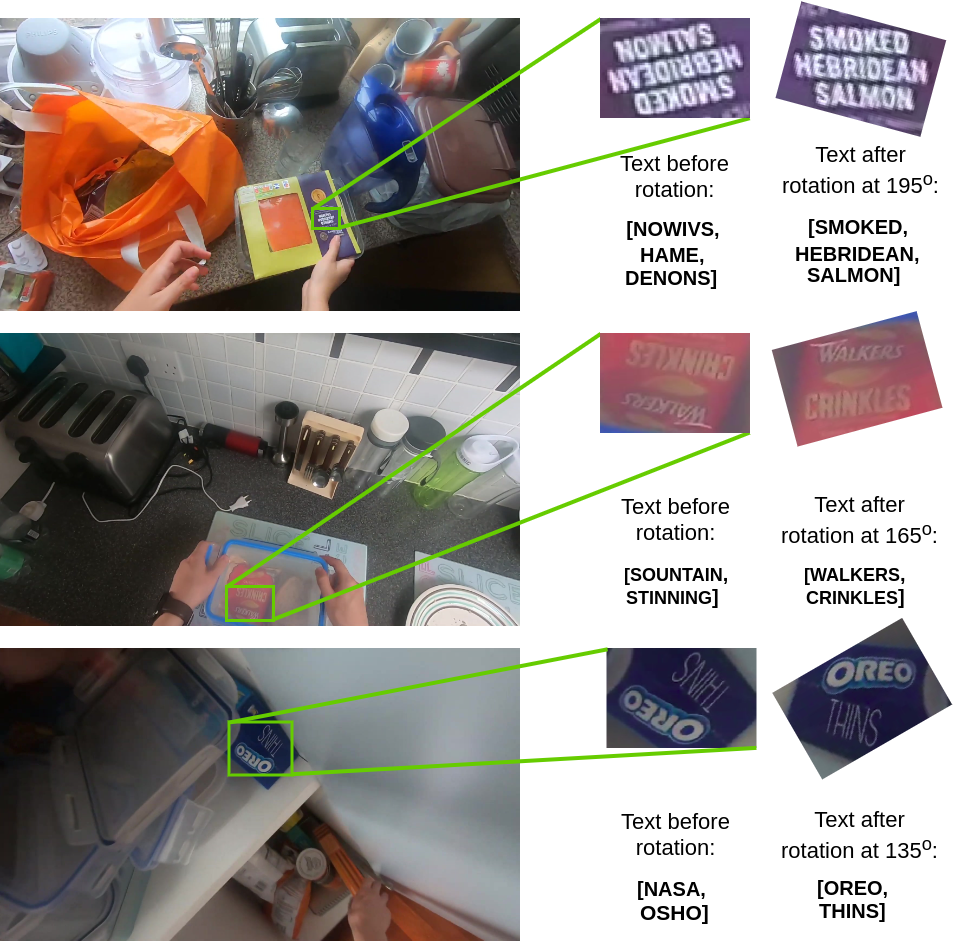}
  \caption{Examples of text recognition by ABINet \cite{abi} with and without the proposed rotation procedure.}
  \label{fig:rotations}
  \vspace{-2mm}
\end{figure}

\noindent\textbf{Results: }
Tab. \ref{tab:results_no_rot} shows text recognition results across all methods with no rotation applied. Tab. \ref{tab:results_rot} shows results for the same methods, when we integrate our proposed rotation pipeline.
Clearly, all methods perform significantly better as part of the rotation framework. In particular, the best performing method (with and without rotation), ABINet \cite{abi}, improves its accuracy from 16.5\% to 49.6\%, and normalised edit distance improves from 0.67 to 0.31. Examples are shown in Fig. \ref{fig:rotations}.
Additionally, the ranking of methods is consistent with and without rotation, which suggests that there is not a method which handles rotation significantly better than the others. This highlights a significant weakness in all current OCR approaches, and shows that rotation should be accounted for in future works.
As expected, the best performing models are more recent and exploit transformers, compared to legacy network designs in the less successful methods.

\begin{table}[t]
\centering
\resizebox{0.47\textwidth}{!}{%
\begin{tabular}{c|ccc}
\hline
Method        & Accuracy $\uparrow$ & Avg. ED $\downarrow$ & Norm. ED $\downarrow$  \\
\hline
ABINet \cite{abi}       & 49.6    & 2.43                  & 0.31                            \\
SAR \cite{sar}          & 47.3    & 2.56                  & 0.32                            \\
SATRN \cite{satrn}         & 45.3     & 2.62                  & 0.34                            \\
RobustScanner \cite{robustscanner} & 42.0    & 2.6                   & 0.33                            \\
SegOCR \cite{mmocr}          & 31.0    & 3.27                  & 0.37                           \\
NRTR \cite{nrtr}          & 23.7    & 3.67                  & 0.48                           \\
TPS \cite{tps}           & 14.7    & 3.89                  & 0.55                            \\
CRNN \cite{crnn}          & 11.6    & 4.22                  & 0.6                             \\
\hline
\end{tabular}
}\caption{Recognition results obtained with the rotation pipeline.}
\label{tab:results_rot}
\vspace{-2mm}
\end{table}

%-------------------------------------------------------------------------
\section{Conclusion}

In this paper, we investigated the ability OCR methods have to recognise text in images from the egocentric EPIC-KITCHENS dataset. We introduced an effective rotate-and-merge procedure which was applied to pre-trained state-of-the-art OCR methods, and demonstrated large improvements against all non-rotation-aware baselines. This highlights a significant oversight of current OCR approaches. We also release our annotated data for future research at: \url{github.com/tobyperrett/epic-text-annotations}

Avenues for future work include exploring rotation as part of model design or training, annotating more text to enable fine-tuning models, as well as combining information from multiple frames to improve recognition accuracy.
The ability to robustly recognise text in egocentric footage will provide a number of benefits, such as the ability to incorporate text into action and object recognition models.

%%%%%%%%% REFERENCES
{\small
\bibliographystyle{ieee_fullname}
\bibliography{main}
}

\end{document}